\documentclass[letterpaper, 10 pt, conference]{ieeeconf}  

\IEEEoverridecommandlockouts                              

\usepackage{graphics} 
\usepackage{epsfig} 
\usepackage{times} 
\usepackage{amsmath} 
\usepackage{amssymb}  
\usepackage{url}
\usepackage{multirow}
\usepackage{booktabs}
\usepackage{subcaption}
\usepackage{siunitx}

\title{\LARGE \bf
	Scene Coordinate and Correspondence Learning for Image-Based Localization
}

\author{Mai Bui$^{1}$, Shadi Albarqouni$^{1}$, Slobodan Ilic$^{1,2}$ and Nassir Navab$^{1,3}$
	\thanks{$^{1}$ Department of Informatics, Technichal University Munich, Germany}%
	\thanks{$^{2}$ Siemens AG, Munich, Germany}%
	\thanks{$^{3}$ Whiting School of Engineering, Johns Hopkins University, USA}%
}

\begin{document}

	\maketitle
	\thispagestyle{empty}
	\pagestyle{empty}

\begin{abstract}
	Scene coordinate regression has become an essential part of current camera re-localization methods. Different versions, such as regression forests and deep learning methods, have been successfully applied to estimate the corresponding camera pose given a single input image. In this work, we propose to regress the scene coordinates pixel-wise for a given RGB image by using deep learning.
	Compared to the recent methods, which usually employ RANSAC to obtain a robust pose estimate from the established point correspondences, we propose to regress confidences of these correspondences, which allows us to immediately discard erroneous predictions and improve the initial pose estimates. Finally, the resulting confidences can be used to score initial pose hypothesis and aid in pose refinement, offering a generalized solution to solve this task.
\end{abstract}
\section{Introduction}
Camera re-localization from a single input image is an important topic for many computer vision applications such as SLAM \cite{mur2015orb}, augmented reality and navigation \cite{lee2016rgb}. Due to rapid camera motion or occlusions, tracking can be lost, making re-localization methods an essential component of such applications. Early methods focus on keyframe \cite{glocker2015real} or descriptor matching by using SIFT or ORB \cite{rublee2011orb} features to obtain point correspondences from which the camera pose could be inferred. However, those methods usually do not perform well under occlusion or in poorly textured environments.
On the other side, machine learning methods have recently shown great capabilities in estimating camera poses from a single image. In particular, regression forests have been employed for a robust pixel-to-scene coordinate prediction. Correspondence samples are then used to obtain multiple pose hypotheses, and a robust pose estimate is found using RANSAC \cite{shotton2013scene,guzman2014multi,cavallari2017fly,valentin2015exploiting}.

Recently, deep learning methods have emerged, mainly focusing on RGB images as input and directly regressing the camera pose, thus offering fast camera pose estimates \cite{kendall2015posenet,kendall2017geometric}. Most of these methods, however, cannot achieve an accuracy similar to the scene coordinate regression approaches. This leads to the assumption that the intermediate step of regressing scene coordinates plays a crucial role in estimating the camera pose for deep learning algorithms and the generalization of those methods. Moreover, RANSAC \cite{fischler1987random} usually plays a vital role in achieving good accuracy in any of the methods focusing on camera relocalization via scene coordinate regression.

In this paper, we introduce a method, which as a first step, densely regresses pixel-wise scene coordinates given an input RGB image using deep learning. In addition, we propose a new form of regularization, that smoothes the regressed coordinates and which can be applied to further improve the regressed coordinates. Thus, a detailed analysis of scene coordinate regression and the influence of different loss functions on the quality of the regressed scene coordinates is conducted. Our main contribution is seen in the second step, where the confidences of the obtained image to scene coordinate correspondences are regressed, based on which erroneous predictions can immediately be discarded, which in turn results in a more robust initial pose hypothesis. In addition, the resulting confidence predictions can be used to optimize the estimated camera poses in a refinement step similar to previous works \cite{brachmann2017dsac}. In contrast to these methods, our approach offers a more general solution by not restricting itself in terms of the optimization function and thresholding, which are typically used to define the inliers and outliers in RANSAC optimizations.


\section{Related Work}

There exists a vast amount of research focusing on the topic of camera pose estimation. The most related to our work can be divided into two categories: correspondence learning and direct regression. The first focuses on descriptor matching to obtain point correspondences from which the camera pose can be inferred, by using either analytical solutions or learning methods. The second performs direct pose regression by using deep learning methods to obtain a pose estimate from a single input image.

\textbf{Correspondence Learning.} Well-known methods working on the topic of camera re-localization have used random forests for correspondence prediction. Here, the forest is trained to predict pixel to 3D coordinate correspondences, from which the camera pose can then be inferred and iteratively refined in a pre-emptive RANSAC optimization \cite{shotton2013scene}. Several extensions and improvements of the method have been proposed, increasing its performance and robustness \cite{valentin2015exploiting,guzman2014multi,cavallari2017fly}. 

On the other hand, due to the recent success of this method, various related methods have used deep learning approaches. 
Inspired by the approach of \cite{shotton2013scene}, Brachmann et al. \cite{brachmann2017dsac} use two convolutional neural networks (CNNs) to predict the pose for an RGB image; the first CNN is used to predict point correspondences and is linked to a second CNN by a differentiable version of RANSAC, they name DSAC. Notably, reinforcement learning is used to obtain a probabilistic selection to enable end-to-end learning of the framework.

Recently, Schmidt et al. \cite{schmidt2017self} automatically find correpondences in RGB-D images by relying on learned feature representations for correspondence estimation using a 3D model of the scene. A fully-convolutional neural network is trained on a contrastive loss to produce pixel-wise descriptors. Here, pixels corresponding to the same model coordinate are mapped together whereas the remaining pixels have dissimilar feature descriptors. Despite the lack of complete guarantee that the descriptors learned from one video can be mapped to the features of another video capturing the same scene, the method showed robustness and generalization. 

Instead of relying on feature representations for point correspondences, Zamir et al. \cite{zamir2016generic} investigate a generic feature representation based on the viewpoint itself, which could be used to retrieve a pose estimate. A siamese network is trained on a multi-task loss including the pose and a matching function. Image patches are extracted and matched according to their poses, and the network is trained to match patches with similar viewpoints. Additionally, it is shown that the resulting models and features generalize well to novel tasks, such as, scene layout or surface normal estimation.

Capturing both geometric and semantic information, Sch\"oneberger et al. \cite{schonberger2018semantic} propose to learn 3D descriptors and include the task of 3D semantic scene completion into a variational encoder-decoder network. By incorporating semantic information, robust descriptors are learned and shown to successfully aid in visual localization tasks even under strong viewpoint and illumination changes. Such additional information has been shown to work well in similar scenarios, such as in localization, when a semantic map of the environment is given as prior information \cite{wang2015lost,mendez2018sedar}.   


\textbf{Direct Regression.} Lately, direct regression approaches have been emerging, which use deep learning methods to regress camera poses from a single image. Mostly CNNs are used in this context to estimate the camera poses \cite{kendall2015posenet,walch2017image,kendall2016modelling,kendall2017geometric}. 
Therefore, Kendall et al. \cite{kendall2015posenet} use a CNN, called PoseNet, to directly predict the six degrees of freedom for camera pose estimation from RGB images. They parameterize rotation by quaternions, which leads to a total of seven parameters to regress for rotation and translation. Although Kendall et al. \cite{kendall2015posenet} achieve reasonable performance for several indoor scenes, this method still shows a significant inaccuracy as compared with the random forest approaches. Therefore, we assume that predicting an intermediate representation such as point correspondences is important for inferring the final pose. However, this method relies only on RGB images, without the need for depth information, which makes it easily applicable in indoor and outdoor settings.

Walch et al. \cite{walch2017image} extend this approach and connect the last feature vector of the neural network to four LSTM units before concatenating the results and feeding this feature vector to the final regression layer. As in \cite{kendall2015posenet}, a pre-trained CNN used for classification is adapted and fine-tuned to enable the regression of the camera pose. By connecting LSTMs and thus correlating the spatial information, the receptive field of each pixel is enlarged substantially, improving the final pose estimation. 

In \cite{kendall2017geometric}, the authors of \cite{kendall2015posenet} extend their method by introducing novel loss functions, which further reduces the gap in accuracy compared to the state-of-the-art methods.
Further, they show that the re-projection error could be used to additionally fine-tune the model and optimize the regression prediction when the depth information is given. 

As a first direct regression approach that achieves comparable accuracy with regard to the scene coordinate regression methods, \cite{valada18icra} proposes a multi-task learning framework. By combining the global and relative pose regression between the image pairs, the authors present a framework for localization and odometry, which shows great improvements in accuracy.

In comparison, our approach is most related to scene coordinate regression methods using deep learning \cite{brachmann2017dsac}. Scene coordinates are densely regressed as opposed to patch-based regression proposed in the state-of-the art methods. Moreover, correspondence confidences are predicted to remove outliers and boost the accuracy of the initial pose hypothesis. Additionally, the resulting confidences can be directly used for hypothesis scoring and pose refinement.


\section{Methodology}
Given an input RGB image $\textbf{I} \in \mathbb{R}^{h \times w \times 3}$ of a scene, where $h$ and $w$ are the image height and width, respectively, our goal is to estimate the corresponding camera pose, given by its orientation $R \in \mathbb{R}^{3 \times 3}$ and position $\textbf{t} \in \mathbb{R}^3$. The camera pose describes the mapping between the camera and the scene coordinates $\textbf{x}$ and $\textbf{X} \in \mathbb{R}^3$ as
\begin{equation}
\textbf{X} = R\textbf{x}+t.
\end{equation}

The relation between the 3D camera coordinates $\textbf{x} \in \mathbb{R}^3$ and the image pixels $\textbf{p}_x \in \mathbb{R}^2$ depends on the camera's focal length $f_x$, $f_y$ and the optical center $c_x$, $c_y$, and is defined as

\begin{equation}
\textbf{p}_x = (\frac{f_xx}{d}+c_x, \frac{f_yy}{d}+c_y)^T,
\end{equation}
with $\textbf{x} = (x,y,d)^T$ being a point in the camera coordinate frame, given by its coordinates $x,y$ and its depth value $d$. 
In case the camera pose is unknown, it can be retrieved given $N_k$ number of correspondences either, if depth information is available, using the Kabsch algorithm from the 3D-3D correspondences between $\textbf{x}$ and $\textbf{X}$ or using the PnP algorithm \cite{lepetit2009epnp} from the 2D-3D correspondences between the image points $\textbf{p}_x$ and $\textbf{X}$. 

\begin{figure*}
	\centering
	\vspace{0.3cm}
	\includegraphics[width=0.95\linewidth]{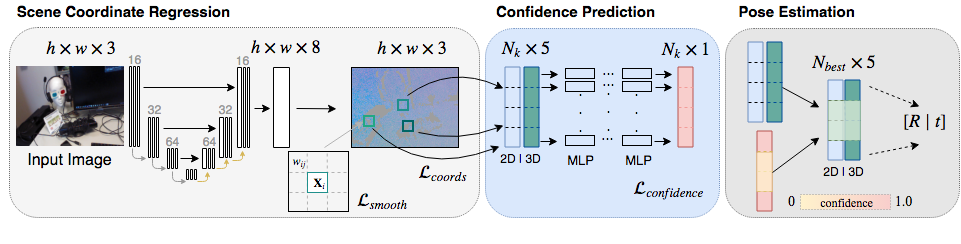}
	\vspace{0.35cm}
	\caption{Outline of our re-localization framework. Scene coordinates are densely regressed for each pixel in the input RGB image. Confidences are predicted for point correspondences and finally used to compute the camera pose estimates.}
	\label{fig:framework}
\end{figure*}

For this aim, our proposed framework consists of three steps: (1) scene coordinate regression, in which we densely predict scene coordinates, and in this context, add a novel regularization, (2) confidence prediction, in which we aim to find accurate correspondences in our coordinate predictions, and (3) pose estimation, in which we employ the aforementioned algorithms to compute the camera pose estimate for the most confident predictions $N_{best}$. An overview of our framework is given in Figure \ref{fig:framework}.

\subsection{Scene Coordinate Regression}
\label{sec:regression}
\textbf{Coordinate regression.} As the first step, our aim is to model the function $\phi:\textbf{I} \rightarrow \hat{\textbf{X}}$, obtaining the predicted scene coordinates $\hat{\textbf{X}}$, as $\hat{\textbf{X}} = \phi(\textbf{I}; \omega)$, where $\omega$ are the model parameters.
Therefore, we compute the ground truth scene coordinates $\textbf{X}$ to train a model and regress the scene coordinates of every pixel, obtaining an $\mathbb{R}^{h \times w \times 3}$ output map. For this purpose, we use the Tukey's Biweight loss function \cite{belagiannis2015robust} to regress the 3D coordinates, given as
\begin{equation}
	\mathcal{L}_{coords} = \frac{1}{h \cdot w} \sum_{i=1}^{h \cdot w} \sum_{s=1}^{S} \rho(r_{i,s})~, 
\end{equation}

\begin{equation}
	\text{~with~} \rho(r_{i,s})= \begin{cases}
	\frac{c^2}{6}[1-(1-(\frac{r_{i,s}}{c})^2)^3],& \text{if } |r_{i,s}| \leq c\\
	\frac{c^2}{6},              & \text{otherwise}
	\end{cases}
\end{equation}

\noindent where $r_{i,s} = X_{i,s}-\hat{X}_{i,s}$ is the residual, $S=3$ 
is the number of coordinates to regress and $\rho(.)$ is Tukey's Biweight function. 
In Tukey's Biweight function the choice of the tuning constant $c$ plays a crucial role
, which is proposed to be chosen according to the median absolute deviation over the residuals assuming a Gaussian distribution. Nevertheless, we propose to choose the parameter $c$ depending on the spatial extent of the current scene, where, after empirical evaluation, we found half of the scenes diameter given in meters to provide better results. 
In case of missing depth values and thus missing ground truth scene coordinates, we omit these pixels during training in order not to negatively influence the network.\\

\noindent \textbf{Coordinate smoothing.} The graph Laplacian regularization, has been successfully applied for image denoising 
on image patches \cite{pang2017graph}; therefore, we consider the scene coordinates in a given neighborhood. Minimizing the graph Laplacian regularizer enables us to smooth the image patches with respect to the given graph. Similarly, we consider the scene coordinates as vertices and compute weights according to the depth value at the corresponding pixel 

\begin{equation}
	w_{ij} = \frac{e^{-|d_i-d_j|}}{\sum_{k \in K, k \neq i} e^{-|d_i-d_k|}},
\end{equation}
where, given a pixel position $i$, we compute weights $w_{ij}$ for each index $j$ in a given neighborhood $K$. In this case, $d_i$ represent the depth value at index $i$. Finally, we obtain the additional smoothing term in our loss function

\begin{equation}
	\mathcal{L}_{smooth} = \sum_{i}^{N_k} \sum_{j \in K} w_{ij} \cdot \|\hat{\textbf{X}}_i-\hat{\textbf{X}}_j\|_2 ,
\end{equation}
where $\hat{\textbf{X}}$ corresponds to the predicted scene coordinates at a given pixel index. This term loosely pushes the surrounding points closer together, given the fact that their depth values are similar; otherwise, a larger difference between points is accepted.

Overall, we train the model using our loss function, described as $\mathcal{L} = \mathcal{L}_{coords} + \mathcal{L}_{smooth}$.
\begin{figure*}[t]
	\centering
	\includegraphics[width=0.98\linewidth]{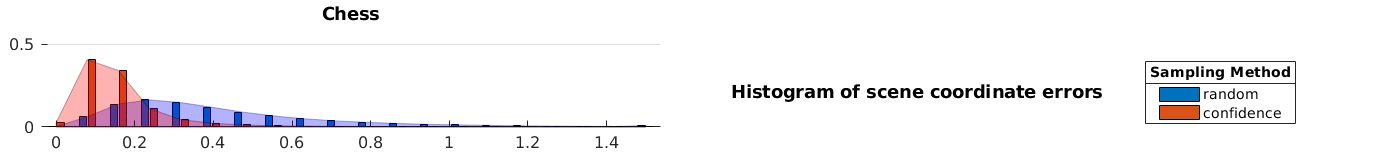}
	\includegraphics[width=0.98\linewidth]{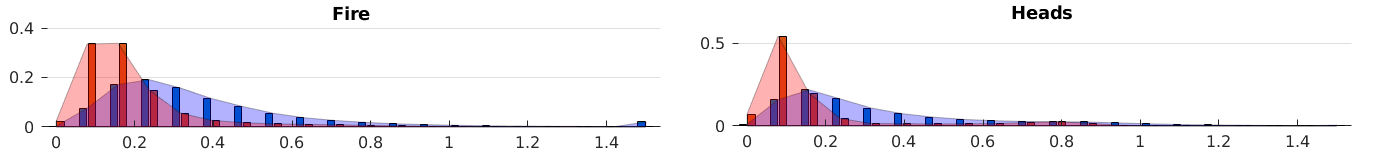}
	\includegraphics[width=0.98\linewidth]{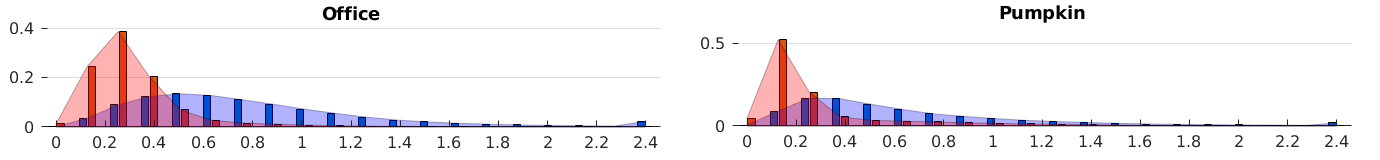}
	\includegraphics[width=0.98\linewidth]{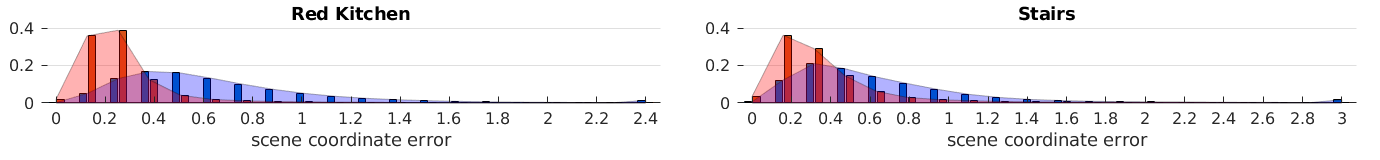}
	\vspace{0.45cm}
	\caption{Normalized histograms of regressed scene coordinate errors used to compute pose estimates on the 7-Scenes. The points are either randomly sampled or, the most confident points out of the samples are used.}
	\label{fig:points_hist}
\end{figure*}
\subsection{Confidence Prediction}
The regressed coordinates can be used to obtain a pose estimate. However, these correspondences usually include a large amount of erroneous correspondences, which is most often solved using RANSAC. \\
%
%
Inspired by \cite{yi2017learning}, which classifies point correspondences between image pairs, we train a neural network to estimate the probabilities of 2D-3D correspondence. 
Instead of solving this as a classification problem, we consider this task as a regression problem. Therefore, we use the model described in the previous section to create probabilities for each scene coordinate prediction and construct the training set $S_{confidence} = \{( \textbf{p}_{\hat{\textbf{X}}_1},\hat{\textbf{X}}_1, \delta_1),..,(\textbf{p}_{\hat{\textbf{X}}_{N_k}}, \hat{\textbf{X}}_{N_k}, \delta_{N_k})\}$, where $\delta_i = e^{-(s \cdot \|\textbf{X}_i - \hat{\textbf{X}_i}\|_2)}$.
\noindent Here, $s$ is used as a scale, so that accurate coordinates are given a high probability. In this step, the objective is, therefore, to compute the function $\Phi: (\textbf{p}_{\hat{\textbf{X}}}, \hat{\textbf{X}}) \rightarrow \delta$, described by the model parameters $\Omega$, so that $\delta =  \Phi((\textbf{p}_{\hat{\textbf{X}}}, \hat{\textbf{X}});\Omega)$. To this end, we feed $N_k$ points containing the image pixel and the predicted scene coordinates to our model. As an output, we obtain a probability for each point according to whether it is likely to be a good correspondence or not. As a loss function, we use the $l_2$ loss to train this model,

\begin{equation}
	\mathcal{L}_{confidence} = \sum_{i}^{N_k} \|\delta_i - \hat{\delta_i} \|_2 .
\end{equation}

The pose predictions can then easily be obtained by sampling the most confident point correspondences, while removing the initial erroneous predictions right away.



\subsection{Pose Estimation and Refinement}
The initial pose hypothesis is refined as a post-processing step. Following previous works \cite{brachmann2017dsac}, $h_p$ pose hypotheses are sampled using $N_k$ number of point correspondences for each. Out of these correspondences, only the $10\%$ points with highest confidence are kept.
However, even though we are able to greatly improve the quality of the point correspondences used to compute pose estimates with this step, the initial randomly selected points might still be highly inaccurate, leading to erroneous predictions included in the confidence sampled subset. To overcome this problem, only one hypothesis out of the $h_p$ is chosen, by scoring each hypothesis using the mean confidence over the probabilities of the correspondences used to compute the pose estimate. 
Then, the best hypothesis is refined by repeatedly sampling $N_k$ randomly selected points and re-running the PnP or Kabsch algorithm, including the additional $10\%$ most confident correspondences in this set. 
\begin{figure*}[t]
	\centering
	\vspace{0.25cm}
	\begin{subfigure}[b]{0.35\textwidth}
		\centering
		\includegraphics[width=\linewidth]{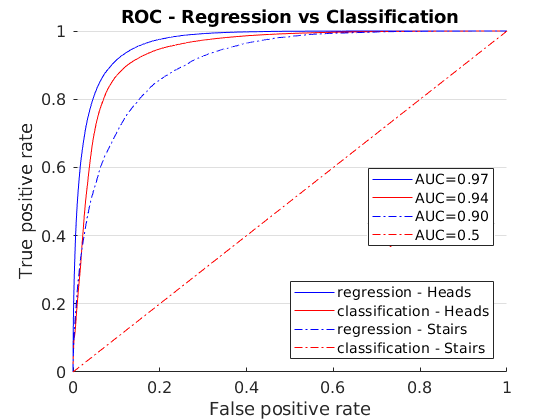}
		\caption{ROC}
		\label{fig:roc}
	\end{subfigure} 
	\begin{subfigure}[b]{0.18\textwidth}
		\centering
		\includegraphics[width=\linewidth]{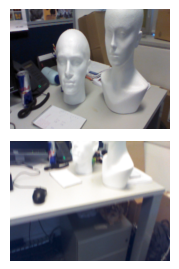}
		\caption{RGB image}
		\label{fig:normal}
	\end{subfigure} 
	\begin{subfigure}[b]{0.18\textwidth}
		\centering
		\includegraphics[width=\linewidth]{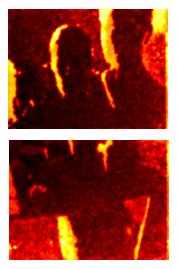}
		\caption{Error map}
		\label{fig:symmetric}
	\end{subfigure}
	\begin{subfigure}[b]{0.22\textwidth}
		\centering
		\includegraphics[width=\linewidth]{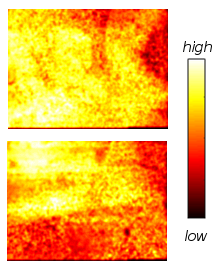}
		\caption{Confidence map}
		\label{fig:rotinv}
	\end{subfigure}
	\vspace{0.45cm}
	\caption{a) ROC comparison between regression and classification for the training images of the \textit{Heads} and \textit{Stairs} scenes, and example images showing b) input RGB, c) corresponding pixel-wise scene coordinate error and d) confidence map predicted by our model.}
	\label{fig:prob_images}
\end{figure*}

\subsection{Implementation Details}
For scene coordinate regression, U-Net is used as the network architecture, as it could easily be used to regress correspondence maps and has been shown to train well on few input images \cite{ronneberger2015u}. Four convolutional layers with pooling layers and dropout applied after each, and four up-sampling layers are used, which gives a final feature map of size $h \times w \times 8$. By applying a last convolutional layer, we obtain the final correspondence map of size $h \times w \times 3$. 

For confidence prediction, we adapt PointNet \cite{qi2017pointnet} to our specific input and output requirements. As our aim is to predict confidences for a subset of point correspondences, we mimic the input for solving the PnP and keeping our final objective in mind feed randomly selected points to PointNet, which has been shown to work well on unstructured data such as point clouds. This way, the network learns to be independent of the order in which the points are fed and could still be used in scenarios where a dense representation of point correspondences is not given.
For the final regression layer, following \cite{yi2017learning}, we first apply a hyperbolic tangent followed by a ReLu activation, so the network can easily predict low confidence for highly inaccurate points. Mostly for evaluation purposes, $s$ is computed
, so that the inlier points have a lower bound probability of 0.75, resulting in $s = 2.8768$.

We set $N_k=500$ out of which only the most confident points are used to estimate a camera pose, resulting in only $N_{best}=50$ points. For pose refinement, we follow the parameter settings of \cite{brachmann2017dsac} and sample $h_p=256$ initial hypotheses and refine the best one for eight iterations on the additional most confident points out of the randomly sampled points.

Both networks are trained separately for 800 epochs with batch size of 20 using RMSprop Optimizer with an initial learning rate of $5 \cdot 10^{-4}$. All experiments were conducted on a Linux-based system with 64-GB RAM and 8-GB NVIDIA GeForce GTX 1080 graphics card and implemented using TensorFlow \cite{abadi2016tensorflow}.
\section{Experiments}
To evaluate our method, we define baseline models, which are described in Section \ref{sec:baseline}, and use the following metrics. For this purpose, inliers and outliers are defined as
$\| \textbf{X} - \hat{\textbf{X}}\|_2 < t_{inlier}$, with $t_{inlier} = 0.1 ~m$ being a common threshold chosen to define inliers \cite{shotton2013scene,valentin2015exploiting,brachmann2017dsac}. Every inlier point is therefore counted as a true positive in our evaluations.
For pose estimation we compute the median rotation, the translation error and the pose accuracy, where a pose is considered correct if the rotation and translation error are below $5^\circ$ and $5~cm$, respectively.

\subsection{Dataset}
\label{sec:dataset}
Our method is evaluated on the publicly available 7-Scenes dataset from Microsoft, which consists of seven indoor scenes with varying volumes ranging from 1 to 18 $m^3$. RGB and depth images of each scene and their corresponding ground truth camera poses are available. For each scene images in the range of 1K to 7K are provided, including very difficult frames due to motion blur, reflecting surfaces and repeating structures, for example, in case of the \textit{Stairs} scene. The images were captured using a Kinect RGB-D sensor and the ground truth poses obtained using KinectFusion. Following the state-of-the-art, we use the training and test sets specified for this dataset. No augmentation as proposed in \cite{wu2017delving,naseer2017deep} was performed and our models were trained individually for each scene.
\subsection{Baseline Models}
\label{sec:baseline}
First, we evaluate each individual component of our model and create the baseline models for comparison. To start with, the first step of our pipeline, the scene coordinate regression is evaluated. To this aim, a model is trained on scene coordinate regression by using different loss functions. Mainly we compare between $l_1$ and the $L_{coords}$ loss as described in section \ref{sec:regression}. In this case, a pose estimate is computed by randomly sampling $N_k$ number of points. Results are given for a single pose estimate. Further, we abbreviate these model as $P_{l1}$ and $P_{tukey}$. Next, to evaluate the scene coordinate regression quality, regularization is added in the form of the introduced smoothness term $L_{smooth}$, corresponding to model $P_{smooth}$.

Further, and more importantly, the second step of our pipeline, the confidence prediction is analyzed. The predicted scene coordinates and associated image points are used to train a model and regress the confidence of each correspondence. For pose estimation, in this case, only the most confident correspondences out of the initial randomly sampled $N_k$ points are kept. We abbreviate this model as $P_{confidence}$.
\setlength{\tabcolsep}{4pt}
\begin{table*}[t]
	\begin{center}
		\caption{Median rotation and translation error on the \textit{Heads} scene for our baseline models. Results are computed using only one pose hypothesis without any further refinement.}
		\label{table:baselines}
		\vspace{0.5cm}
		\resizebox{0.7\textwidth}{!}{%
			\begin{tabular}{lccccc}
				\toprule
				\noalign{\smallskip}
				Model &$h_p$&$P_{l_1}$ &$P_{tukey}$ & $P_{smooth}$ & $P_{confidence}$ \\
				\noalign{\smallskip}
				\midrule
				\noalign{\smallskip}
				\multirow{2}{*}{Kabsch} & $~1~$& $~~18.0^\circ$, $0.29m~~$& $~~8.77^\circ,~0.15m~~$ & $~6.67^\circ,~0.12m~~$ & $~~5.77^\circ,~0.07m~~$\\
				& $256$&$14.7^\circ,~0.25m$ & $6.23^\circ,~0.11m$ & $5.88^\circ,~0.09m$ & $4.86^\circ,~0.06m$\\
				\multirow{2}{*}{PnP} & 1 &$50.3^\circ$, $0.44m$& $41.3^\circ$, $0.44m$& $44.3^\circ,~0.43m$ & $10.6^\circ,~0.18m$\\
				& 256 &$33.7^\circ$, $0.46m$& $25.9^\circ,~0.43m$ & $26.3^\circ,~0.37m$ & $5.09^\circ,~0.10m$\\
				\midrule
				Inliers &1 & 5.1\%& 7.8\% & 11.9\% & 50.7\%\\
				\bottomrule
			\end{tabular}
		}
	\end{center}
\end{table*}
\setlength{\tabcolsep}{1.4pt}
\subsection{Evaluation of Baseline Models}
Each of our models is evaluated, comparing the error between the regressed and ground truth scene coordinates, and the pose error. We compare our models trained on the scene coordinate regression and additionally regularizing the model using the smoothness term. Here, we found a slight improvement in terms of the regressed coordinates as well the pose estimates comparing our models with and without additional regularization. 

Using our proposed confidence prediction, Figure \ref{fig:points_hist} shows that the point errors of the points used to compute a pose estimate significantly decrease, successfully eliminating most of the erroneous predictions and greatly boosting the pose estimation accuracy. As a result, the estimated poses also improve significantly, as seen in Table \ref{table:baselines}. Specifically, the translation error greatly decreases. It should be noted, that only the most confident point correspondences are used to compute a pose estimate for this model. As a result, more accurate poses are obtained using a much smaller number of points
. Additionally, the percentage of inliers in the sampled points used to compute initial pose hypothesis significantly increases. For the evaluation of our proposed confidence prediction to RANSAC, we sampled $h_p$ pose hypotheses and keep the pose hypothesis with the highest inlier score as a final result. Due to the very low amount of inliers, it is difficult to apply RANSAC and obtain satisfactory results without pose refinement or additional confidence prediction.
Furthermore, we analyze the quality of our model's coordinate regression and confidence prediction, where example error and probability maps are shown in Figure \ref{fig:prob_images}. In our case, since we densely regress the scene coordinates, high error values usually correspond to missing depth values and therefore missing ground truth coordinates in the image. Although it seems difficult for the network to accurately predict low confidences in regions with unusually large error, in regions of inlier predictions, the model is able to predict corresponding high confidence.
\subsection{Evaluation of Confidence Prediction}
To assess the quality of regressing correspondence probabilities, a model is trained on simple classification, where a point correspondence could be labeled either as an inlier or an outlier depending on the threshold $t_{inlier}$. The model is trained using cross-entropy loss. As a second step, we train our proposed model, regressing probabilities in the range of $\delta \in [0,1]$ instead and plot the resulting ROC curves as shown in Figure \ref{fig:roc}. As a result, we assess that the performance of regression in this case is more or less equal to classification. However, we do not restrict the model to a specific threshold chosen for the inlier definition, which needs to be adapted for each scene depending on the quality of scene coordinate regression. In comparison, a classification model trained on the challenging \textit{Stairs} scene results in a drop of relative rotation and translation error of $4.4\%$ and $31.5\%$, respectively, because very few inliers were available during training.

\subsection{Comparison to the state-of-the art}
\setlength{\tabcolsep}{4pt}
\begin{table*}[t]
	\begin{center}
		\caption{Median rotation and translation error on the 7-Scenes dataset. Percentages are given for poses below $5^\circ$ and $5cm$ threshold.}
		\label{table:results}
		\vspace{0.0cm}
		\resizebox{\textwidth}{!}{%
			\begin{tabular}{l|c|cccccc|cc}
				\multicolumn{2}{c}{}&\multicolumn{6}{c}{RGB information}&\multicolumn{2}{c}{RGB-D information}\\
				\toprule\noalign{\smallskip}
				Scene & Volume & \vtop{\hbox{\strut PoseNet}\hbox{\strut LSTM \cite{walch2017image}}}& \vtop{\hbox{\strut PoseNet}\hbox{\strut 2017 \cite{kendall2017geometric}}} &
				\vtop{\hbox{\strut VLocNet}\hbox{\strut STL \cite{valada18icra}}} &\vtop{\hbox{\strut VLocNet}\hbox{\strut MTL \cite{valada18icra}}} & DSAC \cite{brachmann2017dsac} & $Ours_{2D-3D}$& $Ours_{3D-3D}$ & SCoRe \cite{shotton2013scene}\\
				\noalign{\smallskip}
				\midrule
				\noalign{\smallskip}
				Chess & $6m^3$ & $5.7^\circ$, $24cm$& $4.8^\circ$, $13cm$ & 
				$1.7^\circ$, $4cm$&$1.7^\circ$, $3cm$ & $\textbf{1.2}^\circ$, $\textbf{2}cm$ & $1.3^\circ$, $3cm$ (83.0\%)& $1.2^\circ$, $3cm$ (85.7\%)& \textbf{92.6}\%\\
				Fire & $2.5m^3$& $11.9^\circ$, $34cm$& $11.3^\circ$, $27cm$&
				$5.3^\circ$, $\textbf{4}cm$ &$4.9^\circ$, $\textbf{4}cm$ & $\textbf{1.5}^\circ$, $\textbf{4}cm$ & $2.9^\circ$, $6cm$ (42.4\%) &$2.7^\circ$, $5cm$ (48.8\%)& \textbf{82.9}\%\\
				Heads & $1m^3$ & $13.7^\circ$, $21cm$& $13^\circ$, $17cm$&
				$6.6^\circ$, $5cm$&$5.0^\circ$, $5cm$ & $\textbf{2.7}^\circ$, $\textbf{3}cm$  & $3.2^\circ$, $4cm$ (59.6\%)& $3.1^\circ$, $3cm$ (\textbf{60.1}\%)& 49.4\%\\
				Office & $7.5m^3$ & $8.0^\circ$, $30cm$& $5.5^\circ$, $19cm$&
				$2.0^\circ$, $4cm$& $\textbf{1.5}^\circ$, $\textbf{3}cm$ & $1.6^\circ$, $4cm$ & $2.1^\circ$, $6cm$ (42.5\%)& $2.0^\circ$, $5cm$ (49.2\%)& \textbf{74.9}\%\\
				Pumpkin & $5m^3$ & $7.0^\circ$, $33cm$& $4.7^\circ$, $26cm$ & 
				$2.3^\circ$, $\textbf{4}cm$ & $\textbf{1.9}^\circ$, $\textbf{4}cm$ &$2.0^\circ$, $5cm$  & $2.9^\circ$, $\textbf{4}cm$ (62.2\%)& $1.3^\circ$, $3cm$ (66.6\%)& \textbf{73.7}\%\\
				Kitchen & $18m^3$ & $8.8^\circ$, $37cm$& $5.3^\circ$, $23cm$ & 
				$2.3^\circ$, $4cm$&$\textbf{1.7}^\circ$, $\textbf{3}cm$&$2.0^\circ$, $5cm$  & $2.7^\circ$, $5cm$ (58.2\%)& $1.3^\circ$, $4cm$ (66.4\%)& \textbf{71.8}\%\\
				Stairs & $7.5m^3$ & $13.7^\circ$, $40cm$& $12.4^\circ$, $35cm$& 
				$6.5^\circ$, $10cm$&$\textbf{5.0}^\circ$, $\textbf{7}cm$ & $33.1^\circ$, $1.17m$ & $6.3^\circ$, $13cm$ (9.9\%)& $6.1^\circ$, $13cm$ (11.6\%)& \textbf{27.8}\%\\
				\midrule
				\multicolumn{2}{c|}{Average}  & $9.8^\circ$, $31.3cm$& $8.1^\circ$, $22.9cm$ & 
				$3.8^\circ$, $5cm$&$\textbf{3.1}^\circ$, $\textbf{4}cm$& $6.3^\circ$, $20cm$ & $\textbf{3.1}^\circ$, $5.8cm$ (51.1\%)& $2.5^\circ$, $5.2cm$ (55.5\%)& \textbf{67.6}\%\\
				\bottomrule
			\end{tabular}
		}
	\end{center}
\end{table*}
\setlength{\tabcolsep}{1.4pt}
Finally, we report the results of our framework using a combination of scene coordinate regression and confidence prediction, described as $P_{confidence}$. We compare our results to the current state-of-the art methods, namely PoseNet \cite{kendall2017geometric}, which directly regresses the camera poses from the RGB input images and refines the trained models by optimizing on the re-projection error. The median rotation and translation errors evaluated on the 7-Scenes dataset can be found in Table \ref{table:results}, where we report the results obtained using PnP ($Ours_{2D-3D}$) as well as, given depth information is available, using Kabsch algorithm ($Ours_{3D-3D}$). Our model does not depend in any way on the algorithm used to compute pose predictions; therefore, we can easily interchange these algorithms without the need to train additional models.

In most cases, we found a significant improvement in pose accuracy compared to \cite{kendall2017geometric} and adaptations of this method \cite{walch2017image}. A recent method, namely VLocNet \cite{valada18icra}, has achieved comparable results to our method, however using direct pose regression. By evaluating the layers up until weights are shared between global pose regression and the odometry stream, they are finally able to achieve an overall median rotation and translation error of $3.1^\circ$, $4.2~cm$ on the 7-Scenes dataset.

In addition, we compare to recent works on scene coordinate regression, \cite{brachmann2017dsac}. Although, there has been a very recent version of this work \cite{brachmann2017learning}, we compare to the earlier version, since its framework is more similar to our approach, keeping the hypothesis scoring CNN in mind. With the exception of the challenging \textit{Stairs} scene, the state-of-the art method shows slightly better accuracy in terms of RGB pose estimation considering each scene individually. On average our method shows good performance compared to the state-of-the art. 

Although our confidence prediction significantly improves the results, the initial scene coordinate regression still seems erroneous, which will be further explored in future work considering optimizations in handling missing depth and thus ground truth scene coordinates as well. Given that the depth information is available, improvements of the accuracy using RGB-D information can easily be obtained since neither the models nor the pose refinement rely on these algorithms. The results including RGB-D information can be seen in Table \ref{table:results}, for which we give a comparison to the state-of-the art scene coordinate regression forest approach \cite{shotton2013scene}. Also, since we only depend on the most confident points, our results were obtained using a smaller number of points. 
Further, RANSAC based optimization, as applied in most state-of-the art methods, could be easily applied to obtain more accurate pose estimates. For evaluation and comparison to current RGB methods, we keep the parameter settings for pose refinement as proposed in \cite{brachmann2017dsac}. 

In terms of computational time, our method is evaluated on an Intel Core i7 4.2 GHz CPU. The only part, which is running on GPU is the evaluation of the networks. To solve the PnP, we use OpenCV's implementation of \cite{lepetit2009epnp}. Our framework 
runs in $1.06~s$, most of which is due to hypothesis sampling ($0.8~s$) and refinement ($0.26~s$).
In comparison, DSAC, implemented in C++, reports a run-time of approximately $1.5~s$, whereas our method, implemented in Python, already performs well.
\section{Conclusion}
	In this work, we present a framework for dense scene coordinate regression in the context of camera re-localization using a single RGB image as the input. When the depth information is available for obtaining the camera coordinates, the corresponding scene coordinates could be regressed and used to obtain a camera pose estimate. We incorporate this information into the network and analyze how the scene coordinate regression can be further optimized using a smoothing term in the loss function. In addition and more importantly, we predict confidences for the resulting image point to scene coordinate correspondences, from which the camera pose can be inferred, thus eliminating most of the outliers in advance and greatly improving the accuracy of the estimated camera poses. As a final step, the resulting confidences can be used to refine the initial pose estimates, which further improves the accuracy of our method.

\bibliographystyle{IEEEtran}

\begin{thebibliography}{10}
	\providecommand{\url}[1]{#1}
	\csname url@rmstyle\endcsname
	\providecommand{\newblock}{\relax}
	\providecommand{\bibinfo}[2]{#2}
	\providecommand\BIBentrySTDinterwordspacing{\spaceskip=0pt\relax}
	\providecommand\BIBentryALTinterwordstretchfactor{4}
	\providecommand\BIBentryALTinterwordspacing{\spaceskip=\fontdimen2\font plus
		\BIBentryALTinterwordstretchfactor\fontdimen3\font minus
		\fontdimen4\font\relax}
	\providecommand\BIBforeignlanguage[2]{{%
			\expandafter\ifx\csname l@#1\endcsname\relax
			\typeout{** WARNING: IEEEtran.bst: No hyphenation pattern has been}%
			\typeout{** loaded for the language `#1'. Using the pattern for}%
			\typeout{** the default language instead.}%
			\else
			\language=\csname l@#1\endcsname
			\fi
			#2}}
	
	\bibitem{mur2015orb}
	R.~Mur-Artal, J.~M.~M. Montiel, and J.~D. Tardos, ``Orb-slam: a versatile and
	accurate monocular slam system,'' \emph{IEEE Transactions on Robotics},
	vol.~31, no.~5, pp. 1147--1163, 2015.
	
	\bibitem{lee2016rgb}
	Y.~H. Lee and G.~Medioni, ``Rgb-d camera based wearable navigation system for
	the visually impaired,'' \emph{Computer Vision and Image Understanding}, vol.
	149, pp. 3--20, 2016.
	
	\bibitem{glocker2015real}
	B.~Glocker, J.~Shotton, A.~Criminisi, and S.~Izadi, ``Real-time rgb-d camera
	relocalization via randomized ferns for keyframe encoding,'' \emph{IEEE
		transactions on visualization and computer graphics}, vol.~21, no.~5, pp.
	571--583, 2015.
	
	\bibitem{rublee2011orb}
	E.~Rublee, V.~Rabaud, K.~Konolige, and G.~Bradski, ``Orb: An efficient
	alternative to sift or surf,'' in \emph{Computer Vision (ICCV), 2011 IEEE
		international conference on}.\hskip 1em plus 0.5em minus 0.4em\relax IEEE,
	2011, pp. 2564--2571.
	
	\bibitem{shotton2013scene}
	J.~Shotton, B.~Glocker, C.~Zach, S.~Izadi, A.~Criminisi, and A.~Fitzgibbon,
	``Scene coordinate regression forests for camera relocalization in rgb-d
	images,'' in \emph{Computer Vision and Pattern Recognition (CVPR), 2013 IEEE
		Conference on}.\hskip 1em plus 0.5em minus 0.4em\relax IEEE, 2013, pp.
	2930--2937.
	
	\bibitem{guzman2014multi}
	A.~Guzman-Rivera, P.~Kohli, B.~Glocker, J.~Shotton, T.~Sharp, A.~Fitzgibbon,
	and S.~Izadi, ``Multi-output learning for camera relocalization,'' in
	\emph{Proceedings of the IEEE Conference on Computer Vision and Pattern
		Recognition}, 2014, pp. 1114--1121.
	
	\bibitem{cavallari2017fly}
	T.~Cavallari, S.~Golodetz, N.~A. Lord, J.~Valentin, L.~Di~Stefano, and P.~H.
	Torr, ``On-the-fly adaptation of regression forests for online camera
	relocalisation,'' in \emph{CVPR}, vol.~2, 2017, p.~3.
	
	\bibitem{valentin2015exploiting}
	J.~Valentin, M.~Nie{\ss}ner, J.~Shotton, A.~Fitzgibbon, S.~Izadi, and P.~H.
	Torr, ``Exploiting uncertainty in regression forests for accurate camera
	relocalization,'' in \emph{Proceedings of the IEEE Conference on Computer
		Vision and Pattern Recognition}, 2015, pp. 4400--4408.
	
	\bibitem{kendall2015posenet}
	A.~Kendall, M.~Grimes, and R.~Cipolla, ``Posenet: A convolutional network for
	real-time 6-dof camera relocalization,'' in \emph{Computer Vision (ICCV),
		2015 IEEE International Conference on}.\hskip 1em plus 0.5em minus
	0.4em\relax IEEE, 2015, pp. 2938--2946.
	
	\bibitem{kendall2017geometric}
	A.~Kendall and R.~Cipolla, ``Geometric loss functions for camera pose
	regression with deep learning,'' in \emph{Proc. CVPR}, vol.~3, 2017, p.~8.
	
	\bibitem{fischler1987random}
	M.~A. Fischler and R.~C. Bolles, ``Random sample consensus: a paradigm for
	model fitting with applications to image analysis and automated
	cartography,'' in \emph{Readings in computer vision}.\hskip 1em plus 0.5em
	minus 0.4em\relax Elsevier, 1987, pp. 726--740.
	
	\bibitem{brachmann2017dsac}
	E.~Brachmann, A.~Krull, S.~Nowozin, J.~Shotton, F.~Michel, S.~Gumhold, and
	C.~Rother, ``Dsac-differentiable ransac for camera localization,'' in
	\emph{IEEE Conference on Computer Vision and Pattern Recognition (CVPR)},
	vol.~3, 2017.
	
	\bibitem{schmidt2017self}
	T.~Schmidt, R.~Newcombe, and D.~Fox, ``Self-supervised visual descriptor
	learning for dense correspondence,'' \emph{IEEE Robotics and Automation
		Letters}, vol.~2, no.~2, pp. 420--427, 2017.
	
	\bibitem{zamir2016generic}
	A.~R. Zamir, T.~Wekel, P.~Agrawal, C.~Wei, J.~Malik, and S.~Savarese, ``Generic
	3d representation via pose estimation and matching,'' in \emph{European
		Conference on Computer Vision}.\hskip 1em plus 0.5em minus 0.4em\relax
	Springer, 2016, pp. 535--553.
	
	\bibitem{schonberger2018semantic}
	J.~L. Sch{\"o}nberger, M.~Pollefeys, A.~Geiger, and T.~Sattler, ``Semantic
	visual localization,'' in \emph{IEEE Conference on Computer Vision and
		Pattern Recognition (CVPR)}.\hskip 1em plus 0.5em minus 0.4em\relax IEEE,
	2018.
	
	\bibitem{wang2015lost}
	S.~Wang, S.~Fidler, and R.~Urtasun, ``Lost shopping! monocular localization in
	large indoor spaces,'' in \emph{Proceedings of the IEEE International
		Conference on Computer Vision}, 2015, pp. 2695--2703.
	
	\bibitem{mendez2018sedar}
	O.~Mendez~Maldonado, S.~Hadfield, N.~Pugeault, and R.~Bowden, ``Sedar--semantic
	detection and ranging: Humans can localise without lidar, can robots?'' in
	\emph{Proceedings of the 2018 IEEE International Conference on Robotics and
		Automation, May 21-25, 2018, Brisbane, Australia}, 2018.
	
	\bibitem{walch2017image}
	F.~Walch, C.~Hazirbas, L.~Leal-Taixe, T.~Sattler, S.~Hilsenbeck, and
	D.~Cremers, ``Image-based localization using lstms for structured feature
	correlation,'' in \emph{Proceedings of IEEE International Conference on
		Computer Vision (ICCV)}, vol.~1, no.~2, 2017, p.~3.
	
	\bibitem{kendall2016modelling}
	A.~Kendall and R.~Cipolla, ``Modelling uncertainty in deep learning for camera
	relocalization,'' in \emph{Robotics and Automation (ICRA), 2016 IEEE
		International Conference on}.\hskip 1em plus 0.5em minus 0.4em\relax IEEE,
	2016, pp. 4762--4769.
	
	\bibitem{valada18icra}
	A.~Valada, N.~Radwan, and W.~Burgard, ``Deep auxiliary learning for visual
	localization and odometry,'' in \emph{International Conference on Robotics
		and Automation (ICRA 2018)}.\hskip 1em plus 0.5em minus 0.4em\relax IEEE,
	2018.
	
	\bibitem{lepetit2009epnp}
	V.~Lepetit, F.~Moreno-Noguer, and P.~Fua, ``Epnp: An accurate o (n) solution to
	the pnp problem,'' \emph{International journal of computer vision}, vol.~81,
	no.~2, p. 155, 2009.
	
	\bibitem{belagiannis2015robust}
	V.~Belagiannis, C.~Rupprecht, G.~Carneiro, and N.~Navab, ``Robust optimization
	for deep regression,'' in \emph{Proceedings of the IEEE International
		Conference on Computer Vision}, 2015, pp. 2830--2838.
	
	\bibitem{pang2017graph}
	J.~Pang and G.~Cheung, ``Graph laplacian regularization for image denoising:
	analysis in the continuous domain,'' \emph{IEEE Transactions on Image
		Processing}, vol.~26, no.~4, pp. 1770--1785, 2017.
	
	\bibitem{yi2017learning}
	K.~M. Yi, E.~Trulls, Y.~Ono, V.~Lepetit, M.~Salzmann, and P.~Fua, ``Learning to
	find good correspondences,'' in \emph{IEEE Conference on Computer Vision and
		Pattern Recognition (CVPR)}.\hskip 1em plus 0.5em minus 0.4em\relax IEEE,
	2018.
	
	\bibitem{ronneberger2015u}
	O.~Ronneberger, P.~Fischer, and T.~Brox, ``U-net: Convolutional networks for
	biomedical image segmentation,'' in \emph{International Conference on Medical
		image computing and computer-assisted intervention}.\hskip 1em plus 0.5em
	minus 0.4em\relax Springer, 2015, pp. 234--241.
	
	\bibitem{qi2017pointnet}
	C.~R. Qi, H.~Su, K.~Mo, and L.~J. Guibas, ``Pointnet: Deep learning on point
	sets for 3d classification and segmentation,'' \emph{Proc. Computer Vision
		and Pattern Recognition (CVPR), IEEE}, vol.~1, no.~2, p.~4, 2017.
	
	\bibitem{abadi2016tensorflow}
	M.~Abadi, A.~Agarwal, P.~Barham, E.~Brevdo, Z.~Chen, C.~Citro, G.~S. Corrado,
	A.~Davis, J.~Dean, M.~Devin, \emph{et~al.}, ``Tensorflow: Large-scale machine
	learning on heterogeneous distributed systems,'' \emph{arXiv preprint
		arXiv:1603.04467}, 2016.
	
	\bibitem{wu2017delving}
	J.~Wu, L.~Ma, and X.~Hu, ``Delving deeper into convolutional neural networks
	for camera relocalization,'' in \emph{Robotics and Automation (ICRA), 2017
		IEEE International Conference on}.\hskip 1em plus 0.5em minus 0.4em\relax
	IEEE, 2017, pp. 5644--5651.
	
	\bibitem{naseer2017deep}
	T.~Naseer and W.~Burgard, ``Deep regression for monocular camera-based 6-dof
	global localization in outdoor environments,'' in \emph{Intelligent Robots
		and Systems (IROS), 2017 IEEE/RSJ International Conference on}.\hskip 1em
	plus 0.5em minus 0.4em\relax IEEE, 2017, pp. 1525--1530.
	
	\bibitem{brachmann2017learning}
	E.~Brachmann and C.~Rother, ``Learning less is more-6d camera localization via
	3d surface regression,'' in \emph{IEEE Conference on Computer Vision and
		Pattern Recognition (CVPR)}.\hskip 1em plus 0.5em minus 0.4em\relax IEEE,
	2018.
	
\end{thebibliography}

\end{document}